# An ERP Study of Recursive Possessive Parsing in ASD Children and Its Cognitive Neuro Mechanisms

Chenxi Fu[a], Xiaoyi Wang[a], Ziman Zhuang[b], Caimei Yang[a*]

[a] *School of Foreign Languages, Soochow University,* [b] *Ulink High School of Suzhou Industrial Park*

***Corresponding author**: E-mail address:cmyang@suda.edu.cn(C. Yang)

**Abstract**

Recursive structures are a core property of human language, yet little is known about how children with autism spectrum disorder (ASD) process complex recursion. This ERP study investigated the online processing of two-level recursive possessive structures in Mandarin-speaking children with ASD (n = 12) compared to typically developing (TD) peers (n = 12) using a sentence-picture matching paradigm. ERPs were analyzed for P200 (150–250 ms), N400 (300–500 ms), and P600 (500–1000 ms). Results showed that ASD children exhibited significantly reduced P200 amplitudes and failed to show the typical posterior grammaticality effect, indicating atypical early perceptual processing. No robust N400 violation effect was observed in either group, confirming the mismatch was not a semantic anomaly; however, ASD children showed a reversed anterior effect and an attenuated posterior effect. For the P600, ASD children had significantly reduced amplitudes, no posterior grammaticality effect, and a trend toward delayed latency, reflecting a core deficit in syntactic reanalysis. These findings demonstrate that while lexical-semantic processing is relatively preserved in ASD, the online syntactic computation required for recursion is severely impaired, supporting modular dissociation accounts of language in autism.

## 1. Introduction

Recursive structures have long been regarded as a core property of human language, providing the fundamental mechanism by which a finite set of grammatical rules can generate an infinite number of linguistic expressions (Chomsky, 1957). This property has been central to theoretical linguistics for decades, with some scholars even proposing that recursion constitutes the only uniquely human component of the faculty of language in the narrow sense (Hauser et al., 2002). Within this theoretical context, understanding how children acquire and process recursive structures has become a key issue in language acquisition research. Among the various types of recursion, two-level recursive possessive structures (e.g., *the lamb's monkey's airplane*) have received particular attention because their clear hierarchical embedding relationships provide an ideal window for investigating children's syntactic processing abilities (Pérez-Leroux et al., 2012; Roeper, 2011; Terunuma et al., 2017; Li et al., 2020).

While previous studies have made significant progress in documenting the developmental trajectory of recursive possessives in typically developing (TD) children (Fu et al., 2025; Shi et al., 2019; Chien & Chen, 2024), far less is known about how children with autism spectrum disorder (ASD) process such structures. ASD is a neurodevelopmental disorder characterized by impairments in social communication and interaction, alongside restricted and repetitive patterns of behavior (Li & Huang, 2016). Language difficulties are among the most common concerns in this population, often serving as the initial reason for clinical referral. However, existing research on language comprehension in ASD has predominantly focused on lexical processing or simple sentence structures, with findings suggesting that while vocabulary knowledge may be relatively preserved (Tager-Flusberg et al., 1990; Li, 2009), sentence-level comprehension—particularly for syntactically complex structures—remains a challenge (Eigsti et al., 2007; Stockbridge et al., 2014; Peristeri et al., 2024; Cano-Villagrasa et al., 2025).

Critically, despite the theoretical importance of recursion and the clinical relevance of language deficits in ASD, there is a notable scarcity of research examining how Mandarin-speaking children with ASD process recursive structures. Most previous studies have relied on behavioral measures, which capture only offline outcomes and cannot reveal the real-time cognitive and neural dynamics underlying syntactic processing. Moreover, the few electrophysiological studies that exist have largely focused on semantic integration (Coderre et al., 2017; DiStefano et al., 2019), leaving the neural mechanisms of syntactic recursion largely unexplored. This gap is particularly significant because recursion imposes high demands on hierarchical structure building and working memory (Zhang & Zan, 2007; Liu et al., 2019), making it a sensitive probe for detecting subtle processing atypicalities that may not be evident from behavioral accuracy alone.

To address this gap, the present study aims to investigate the online processing of two-level recursive possessive structures in Mandarin-speaking children with ASD, compared to age-matched TD children, using event-related potential (ERP) techniques. ERP offers high temporal resolution, making it particularly suitable for examining the time course of syntactic processing and for identifying the specific stages at which processing difficulties may arise (Li & Huang, 2016). By adopting a sentence-picture matching paradigm, we aim to dissociate the contributions of semantic and syntactic processing through the analysis of three key ERP components: the P200 (reflecting early perceptual and attentional processing), the N400 (associated with semantic integration), and the P600 (linked to syntactic reanalysis and integration).

Based on previous theoretical accounts and empirical findings, we formulate the following research hypotheses. First, if ASD children's syntactic processing is impaired at an early stage, we would expect reduced or atypical P200 effects, particularly in posterior language-related regions (Cheng et al., 2025). Second, given evidence for relatively preserved lexical-semantic knowledge in high-functioning ASD (Coderre et al., 2017; DiStefano et al., 2019), we hypothesize that N400 responses may not differ significantly between groups in terms of latency, although spatial distribution may be atypical. Third, if ASD children face core difficulties in hierarchical syntactic integration, we predict attenuated P600 effects, reflecting reduced efficiency in syntactic reanalysis (Marquez-Garcia et al., 2022; Pijnacker et al., 2010). Finally, we anticipate that any group differences in neural responses will be accompanied by preserved behavioral accuracy but prolonged reaction times in the ASD group, reflecting increased cognitive effort. Accordingly, this study addresses the following research questions: (1) Are there differences between children with ASD and TD children in processing recursive possessive structures? If differences exist, what are the neural signatures of these differences, as reflected in the P200, N400, and P600 components? (2) What language-related ERP components are involved in processing mismatched recursive possessives in ASD children? (3) What parsing strategies do ASD children employ when dealing with recursive possessive structures?

The overarching aim of this study is to elucidate the cognitive neural mechanisms underlying recursive syntactic processing in children with ASD, thereby contributing to theoretical debates on modularity and compensation in language processing, and providing a neural-level foundation for developing targeted language interventions. By using high-temporal-resolution electrophysiological recordings, we seek to move beyond traditional outcome-based assessments to capture the real-time processing dynamics that characterize the strengths and challenges of language comprehension in autism.

## 2. Methods

### 2.1 Participants

A total of 27 children with autism spectrum disorder (ASD) aged 6 to 11 years ($M$ = 7.32, $SD$ = 2.27) were recruited from the **Suzhou Gusu Lehang Youkang Chinese Medicine Polyclinic** and a local special education school. Fifteen participants were excluded from the final analysis due to excessive EEG artifacts or failure to complete the experimental session. Sixteen typically developing (TD) children aged 6 to 7 years ($M$ = 6.72, $SD$ = 0.28) were recruited from local primary schools and kindergartens, of whom 12 provided valid data. The final sample thus consisted of 12 children with ASD and 12 TD children.

All children with ASD had received a formal medical diagnosis according to the =*Diagnostic and Statistical Manual of Mental Disorders*, 5th Edition (*DSM-5*) criteria and possessed a clinical diagnosis certificate. A three-step screening procedure was implemented to ensure that only children with ASD were included. First, experimenters consulted with parents and special education teachers to identify candidates with ASD who had no history of epilepsy or other neurological disorders and no significant behavioral problems. Second, semantic competence was assessed using the Peabody Picture Vocabulary Test (PPVT), a widely used measure in autism research (Coderre et al., 2017; O'Rourke & Coderre, 2021). Children with ASD who demonstrated satisfactory semantic abilities proceeded to the next stage. Third, intellectual ability was assessed using the Chinese Revised Wechsler Intelligence Scale for Children (WISC-CR), which provides measures of verbal IQ, performance IQ, and full-scale IQ. Only children meeting the criteria for high-functioning ASD were included in the final sample. TD children were recruited through a similarly systematic procedure. Parents were interviewed to exclude any children with a history of neurodevelopmental disorders, including autism, epilepsy, or any neurological diseases. Eligible children then underwent the WISC-CR to assess full-scale IQ and verbal IQ. Additionally, the principal investigator interacted with each TD child prior to the experiment to observe for any atypical symptoms; children exhibiting any such symptoms were excluded from the study.

All participating children had normal or corrected-to-normal hearing and vision. Informed consent was obtained from the parents or guardians of all participants. The two groups were matched in terms of chronological age. Independent-samples $t$-tests revealed significant group differences in full-scale IQ ($t(26) = 4.18$, $p < .001$, Cohen's $d = 1.58$) and verbal IQ ($t(26) = 3.56$, $p < .001$, Cohen's $d = 1.35$), with the ASD group showing lower scores. Detailed participant information is presented in Table 1.

Table 1 Information of Participants

| | Number | Average Age | Verbal-IQ (PPVT score) | IQ |
|---|---|---|---|---|
| TD | 12 | 6.72 (0.28) | 113.00 (9.91) | 107.62 (7.85) |
| ASD | 12 | 7.32 (2.27) | 99.21 (10.58) | 92.93 (10.56) |

**2.2 Stimuli**

The experiment employed a sentence-picture matching paradigm. Stimuli consisted of 70 two-level recursive possessive sentences (10 additional sentences for practice) and corresponding visual pictures depicting possessive relations. All objects and animals in the stimuli were selected from the CHILDES Mandarin Zhou Narratives Corpus to ensure familiarity and ease of comprehension for preschool-aged children. Syllable count (all two syllables) and word frequency were balanced across conditions.

Visual stimuli were 70 full-color digitized photographs (10 for practice) depicting possessive relations. Each picture contained two animate characters (e.g., a bird, a chicken) and one inanimate object (e.g., a balloon, a pencil). Pictures were presented in the center of a computer monitor against a white background, with a viewing distance of approximately 60 cm. A pilot study with adults established that all pictures were labeled consistently (100% accuracy). The same set of pictures was used for both match and mismatch conditions, with each picture appearing twice across the randomized order: once followed by a congruent sentence and once followed by an incongruent sentence. The pictures were created using Canva and followed four design principles: (1) possessive relationships were denoted by a red line connecting the two animal figures, explained to participants prior to the experiment; (2) "possessor" and "possessee" were visually distinguished; (3) no extraneous elements were present that could distract from the narrative; (4) all characters and objects were depicted in a uniform artistic style to ensure visual coherence.

Seventy sentences followed the structure *Possessor de Possessee de object* like example 1, and an additional 70 sentences followed the reversed structure *Possessee de Possessor de object* like example 2, serving as mismatched stimuli. All nouns were disyllabic and selected from the CHILDES Mandarin Zhou Narratives Corpus. The two conditions used identical lexical items, differing only in the order of possessors. To control for animacy effects (Corrigan, 1988), all noun phrases were animate (animal characters), ensuring that neither noun was more likely to be interpreted as the agent based on animacy.

Auditory stimuli were generated using a text-to-speech synthesizer in CapCut with a "broadcasting" style. All words were digitally produced at a sampling rate of 44.1 kHz (16 bit; mono). Each character had a duration of 1000 ms. Sentences were presented auditorily at approximately 65 dB SPL via two loudspeakers positioned 30 degrees to the left and right of midline in front of the participant.

Prior to the experiment, 20 adults not involved in the study evaluated the plausibility of the image-sentence pairs using a 5-point scale (1 = completely unreasonable, 5 = very reasonable). Sentences with a mean plausibility rating below 3 were excluded from further analysis.

Example 1.

zhèshì xiǎoyáng de xiǎohóu de fēi jī

This is sheep-Poss monkey-Poss plane

This is sheep's monkey's plane.

Example 2.

zhèshì xiǎohóu de xiǎoyáng de fēi jī

This is monkey-Poss sheep-Poss plane

This is monkey's sheep's plane.

**2.3 Procedure**

The Match-Mismatch paradigm is a widely employed event-related potential (ERP) experimental design, primarily used to investigate the cognitive processing of sensory stimuli and semantic integration within the brain (Cantiani et al., 2016). By comparing the ERP waveforms elicited under matching and mismatching conditions, this paradigm provides insights into the neural mechanisms underlying consistency processing. In a typical experiment, participants are initially presented with a visual stimulus, such as an image, followed by an auditory stimulus. The matching condition involves congruence between the visual and auditory stimuli (e.g., an image of a dog followed by the word "dog"), while the mismatching condition introduces semantic incongruence (e.g., the same image followed by the word "cat"). Participants are not required to provide an active behavioral response but instead passively view the images and listen to the words. This passive design makes the paradigm particularly suitable for studying populations with limited or no verbal abilities, such as non-verbal or minimally verbal children with autism. The paradigm's utility lies in its ability to reveal differences in neural processing at various stages, from basic sensory perception to higher-order semantic integration, offering valuable insights into the neurocognitive profiles of these participants.

The experiment was conducted in a quiet classroom. Each child was tested individually, and two trained examiners completed the test work together. The parents were seated next to the child, but they could not see the images or hear the sentences, and the child could not see their faces. Parents were instructed to sit still and to remain silent. Before the experiment, the experimenter introduced the items and animal images that would appear in the later experimental stage to the subjects.

During the EEG experiment, children performed sentence-picture matching tasks. The picture-sentence matching task was presented by means of E-prime 3.0(Psychology Software Tools, Inc), paired plot images were displayed in front of the children. A sentence was then fed into the sound equipment. The instruction of formal experiment follows: "Hello. You see there is a computer here. A little robot lives inside it. Sometimes it tells lies. You need to look at the pictures and listen to what the robot says. If the robot says something that matches the item the finger is pointing to in the picture, then press the green tick on the keyboard. If the robot says something different from what the finger is pointing to, then press the red cross on the keyboard. Please listen and look carefully and have fun with the game.” After the subjects are ready, press "Q" to enter the experiment. After the practice stage is over, press "Q" to enter the experiment stage.

During the formal experiment, the computer will present pre-recorded sentences featuring a two-level recursive possessive structure (refer to session 3.2.3 for specifics). These sentences will either correspond to or deviate from the displayed images.

Prior to the formal experiment, a practice session consisting of 10 trials will be conducted. The procedure for the practice session mirrors that of the formal experiment. Only children who demonstrate the ability to accurately judge the correspondence between sentences and images during the practice session, achieving a minimum accuracy rate of 75%, will proceed to the formal experiment. The formal experimental session was divided into two blocks, each containing

60 sentences in a pseudorandomized order. The pictures were randomly distributed among the blocks. Each block lasted about 10 min. A break was included between blocks and additional breaks, if necessary. In total, one complete experimental session did not exceed 25 min.

All participants must have normal hearing and vision, with vision correctable to normal if necessary. Informed consent has been obtained from the parents or guardians of all child participants.

At the beginning of each test, a red gaze point "+" will first appear in the center of the screen, with a display time of 500ms. Then, a conditional sentence will be presented, with a display time of 5000ms. Only after the listening to the sentence, can the participants press the button to answer. After the interface is judged, it is an empty screen that disappears after the main test button is pressed. Then, a red gaze point "+" appears again on the screen for 500ms, and then the next test is played (See Figure 1).

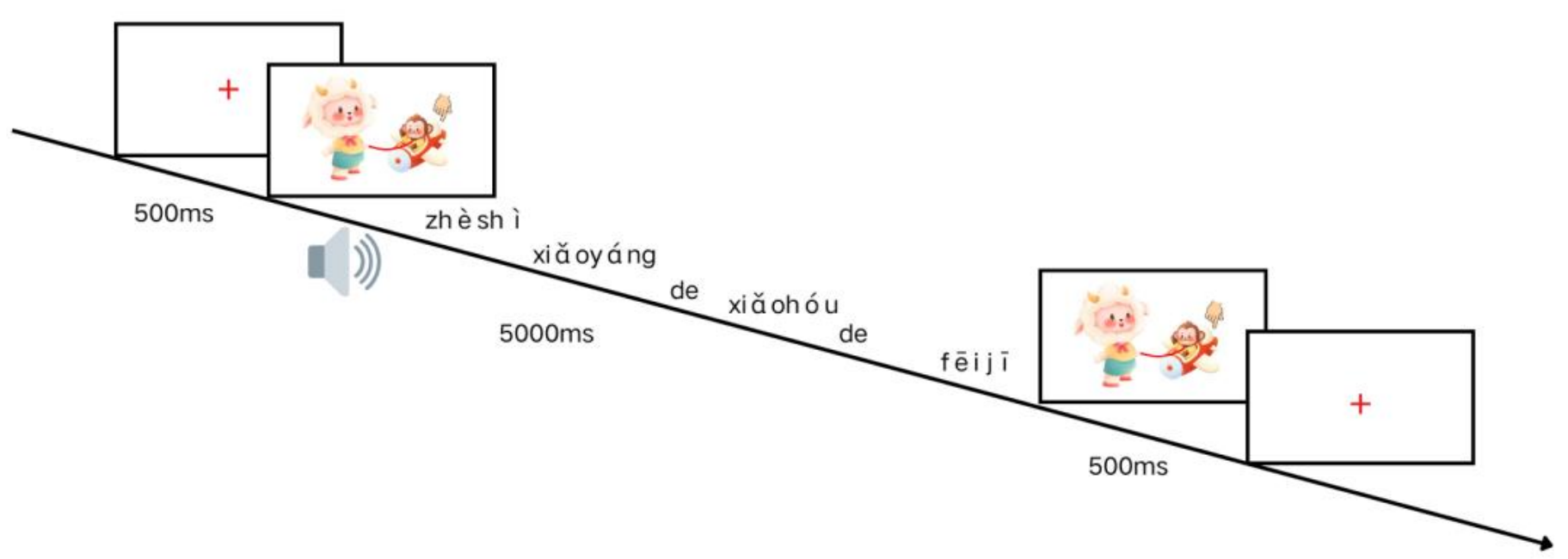

Figure 1. Experimental Procedure

## 2.4 EEG Recording and Preprocessing

Electroencephalogram (EEG) data were continuously recorded using a Smarting Streamer system (mBrainTrain) with 32 scalp electrodes arranged according to the 10–20 International System of Electrode Placement. The electrode distribution was as follows: prefrontal (FP1, FP2); frontal (F7, F3, Fz, F4, F8); frontocentral (FC5, FC1, FCz, FC2, FC6, FT9, FT10); temporal (T7,

T8); central (C3, Cz, C4); centroparietal (CP5, CP1, CP2, CP6); temporoparietal (TP9, TP10); parietal (P7, P3, Pz, P4, P8); and occipital (O1, O2). The ground electrode was placed on the scalp between FCz and Fz. All channels were referenced online to the average of the two mastoid electrodes (TP9 and TP10). Electrode impedances were maintained below 5–10 kΩ. The EEG signal was digitized at a sampling rate of 250 Hz.

Offline preprocessing was performed using EEGLAB and ERPLAB toolboxes (Delorme et al., 2005) running in MATLAB (R Development Core Team, 2013). The data were first re-referenced to the average of TP9 and TP10. A bandpass filter was applied with a high-pass cutoff of 0.1 Hz and a low-pass cutoff of 40 Hz. Continuous EEG data were segmented into epochs ranging from 200 ms before stimulus onset to 1000 ms after stimulus onset. Artifact rejection was performed using automatic channel rejection and manual inspection; epochs containing excessive eye blinks, eye movements, or muscle activity were excluded. Independent component analysis (ICA) was used to identify and remove residual ocular and muscular artifacts (Delorme et al., 2005). Data were then re-referenced to an average reference. For each participant, ERPs were averaged separately for match and mismatch conditions.

### 2.5 Statistical Analysis

Statistical analyses were conducted using R software (R Core Team, 2013). To examine the processing of recursive possessive structures across groups and conditions, we focused on three time windows based on prior literature and visual inspection of the grand-averaged waveforms: P200 (150–250 ms), N400 (300–500 ms), and P600 (500–1000 ms). Sixteen electrodes were selected for analysis based on previous research (Lau & Liao, 2018), divided into four regions of interest: left anterior (F7, F3, FT7, FC3), right anterior (F4, F8, FC4, FT8), left posterior (TP7, CP3, P7, P3), and right posterior (CP4, TP8, P4, P8).

For each time window, mean amplitude data were analyzed using a linear mixed model (LMM) with the following fixed factors: Group (ASD, TD), Grammaticality (match, mismatch), Anteriority (anterior, posterior), and Hemisphere (left, right). Subject was included as a random intercept. Models were fitted using restricted maximum likelihood (REML) with the Satterthwaite method for degrees of freedom estimation. All significant main effects and interactions involving Group and Grammaticality are reported.

For behavioral data (accuracy and reaction time), repeated-measures analyses of variance (ANOVAs) were conducted with Group as a between-subjects factor and Grammaticality as a within-subjects factor. Post-hoc comparisons were performed using Bonferroni-corrected $t$-tests where appropriate.

Latency analyses were conducted for the N400 and P600 components using peak detection methods on difference waves (mismatch minus match) averaged across fronto-central electrodes (Fz, FCz, Cz). Participants without a clear peak exceeding a pre-defined threshold (N400: $> -1$ μV; P600: $> 1$ μV) were excluded from latency analyses. Group differences in peak latency were assessed using independent-samples $t$-tests and Mann-Whitney $U$ tests. Effect sizes (Cohen's $d$) are reported for all comparisons.

## 3. Electrophysiological Results

To investigate the neural mechanisms underlying the processing of two-level recursive possessive structures, we analyzed event-related potentials (ERPs) locked to the onset of the critical word (the second possessor) in three time windows of interest: P200 (150–250 ms), N400 (300–500 ms), and P600 (500–1000 ms). Based on previous research (Lau & Liao, 2018), 16 electrodes were selected and divided into four regions: left anterior (F7, F3, FT7, FC3), right anterior (F4, F8, FC4, FT8), left posterior (TP7, CP3, P7, P3), and right posterior (CP4, TP8, P4, P8). For each time window, mean amplitude data were analyzed using a linear mixed model (LMM) with Group (ASD, TD), Grammaticality (match, mismatch), Anteriority (anterior, posterior), and Hemisphere (left, right) as fixed factors, and Subject as a random intercept. Figures 2 and 3 present the overall average waveforms for the TD and ASD groups, respectively.

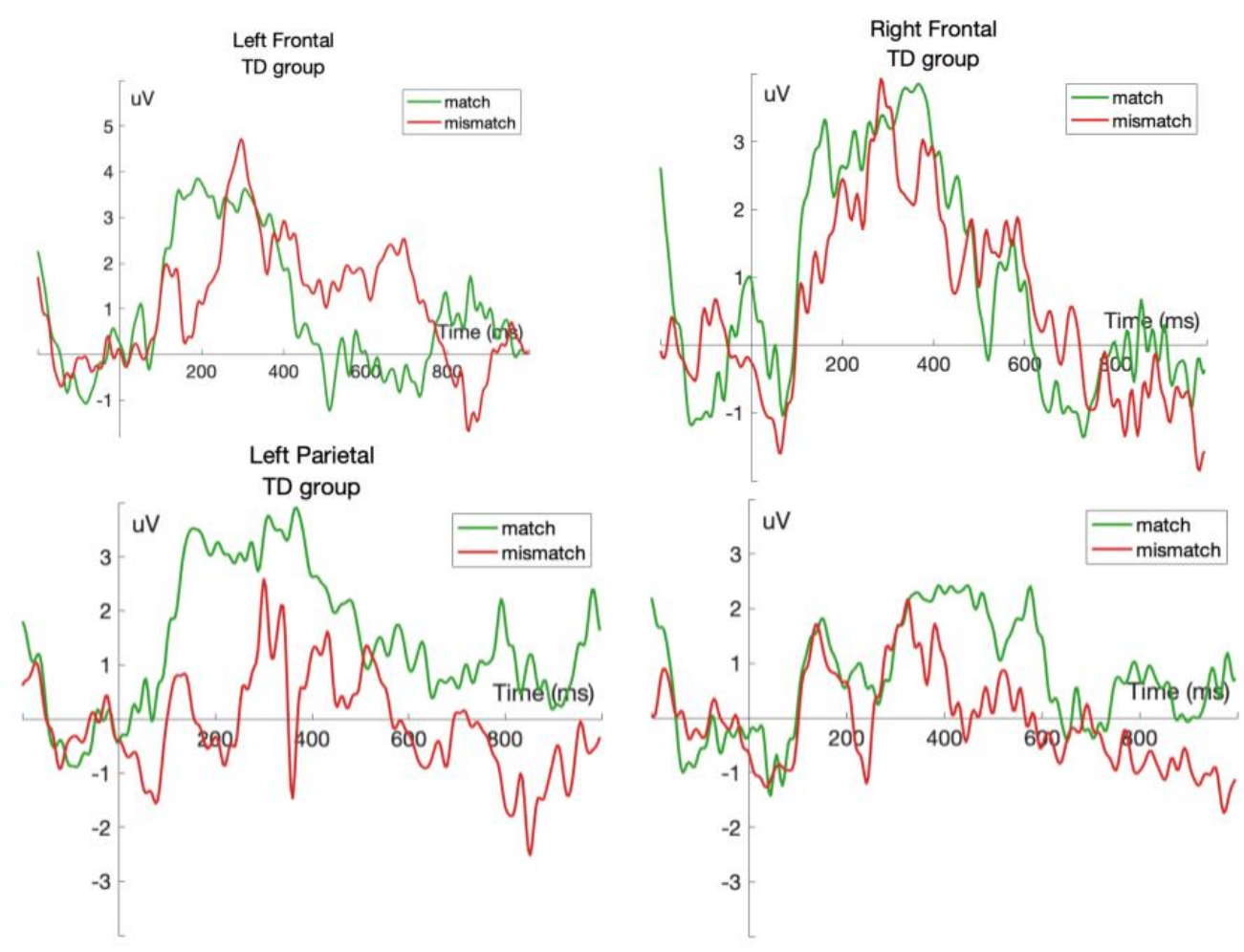

Figure 2. The Overall Average Waveform Diagrams Induced by TD Children's Match and Mismatch types

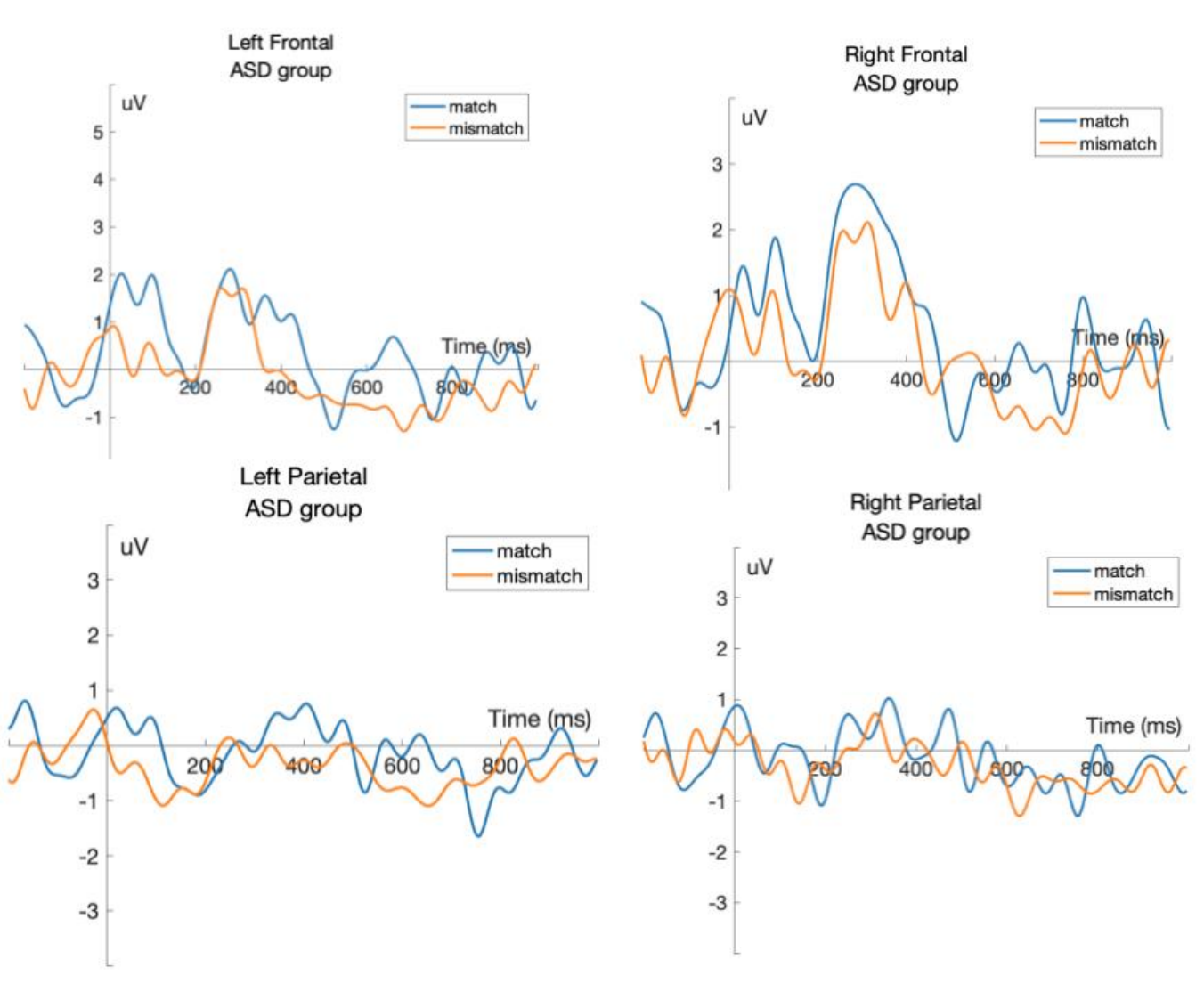

Figure 3. The Overall Average Waveform Diagrams Induced by ASD Children's Match and Mismatch types

**3.1 P200 (150–250 ms)**

The LMM on the P200 mean amplitude revealed a significant main effect of Group (β = –2.0458, SE = 0.4942, t(736.12) = –4.139, p < .001), with the ASD group showing significantly lower overall amplitudes than the TD group. A significant Grammaticality × Anteriority interaction was observed (β = –2.2086, SE = 0.7581, t(734.49) = –2.913, p < .005), indicating that the effect of grammatical processing was modulated by brain region. Furthermore, a significant three-way interaction of Group × Grammaticality × Anteriority emerged (β = 2.4072, SE = 1.1015, t(734.49) = 2.185, p < .05), suggesting fundamental differences in the spatial distribution of early grammatical processing effects between the two groups.

Simple effects analyses within the TD group showed a typical regional specificity: in posterior regions, amplitudes for matching conditions were significantly higher than for mismatching conditions (mean difference = 1.64 μV, $p < .001$), whereas no significant difference was observed in anterior regions (mean difference = 0.452 μV, $p = .183$). In contrast, the ASD group exhibited no significant amplitude differences between conditions in either anterior (mean difference = 0.374 μV, $p = .295$) or posterior (mean difference = 0.234 μV, $p = .59$) regions. Between-group comparisons further revealed that, in the posterior region under mismatching conditions, the difference between groups was not significant (mean difference = 0.37 μV, $p = .395$, Cohen's $d = 0.20$), whereas in the other three conditions, TD children showed significantly higher amplitudes than ASD children (all $p < .001$). The scalp topography (Figure 4) visually confirmed that the grammaticality effect in TD children was concentrated in posterior regions, whereas no such specific pattern was present in the ASD group.

Latency analysis of the P200 component, based on difference waves (mismatch minus match) at fronto-central electrodes (Fz, FCz, Cz), revealed no significant group difference. The mean peak latency was 199.2 ms (SD = 46.2 ms) for the TD group and 206.0 ms (SD = 28.3 ms) for the ASD group ($t(9) = -0.30$, $p = .770$; Cohen's $d = -0.18$).

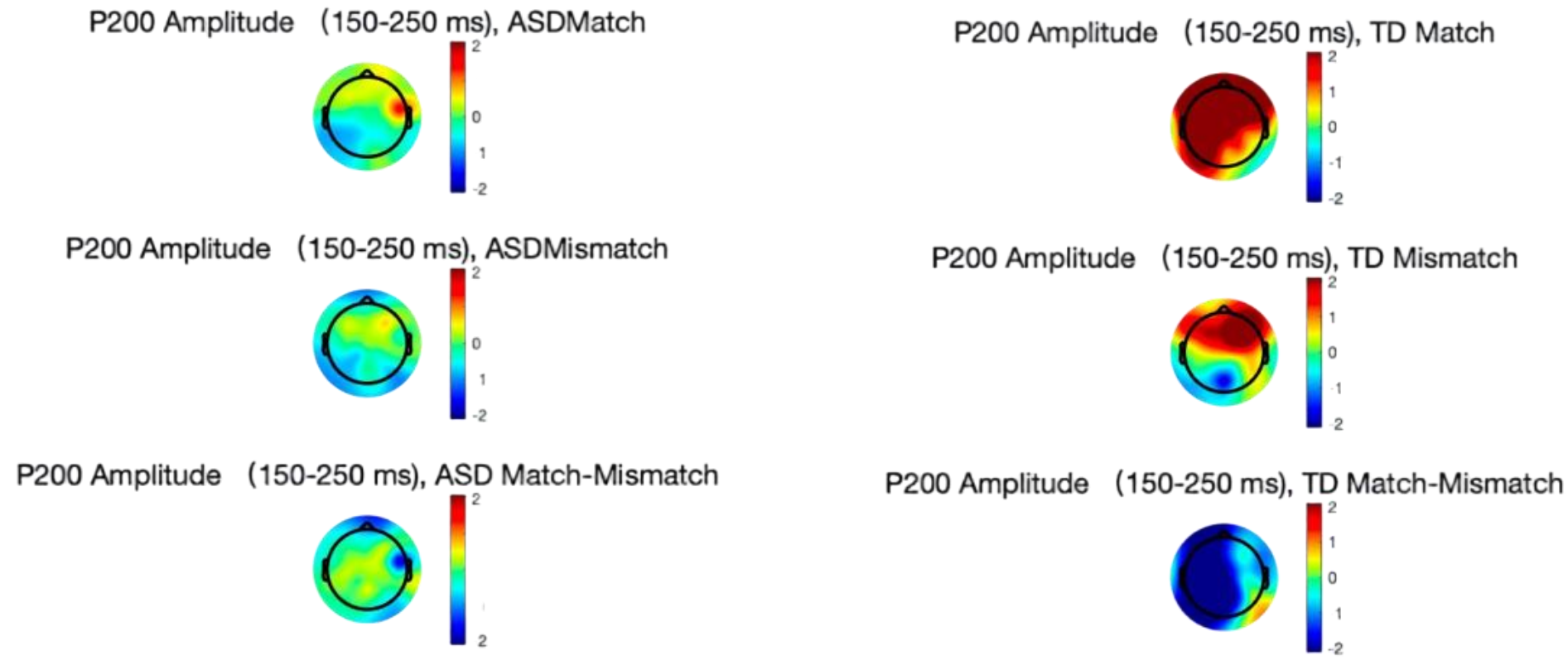

Figure 4. Topography (150ms-250ms) in ASD and TD group

**3.2 N400 (300–500 ms)**

Analysis of the N400 time window revealed a significant three-way interaction of Group × Grammaticality × Anteriority ($\beta = 1.994$, $SE = 1.000$, $t(773.87) = 1.995$, $p = .046$). A significant Grammaticality × Anteriority interaction was also observed ($\beta = -2.118$, $SE = 0.741$, $t(773.87) = -2.857$, $p = .004$). The main effect of Group was significant ($\beta = -2.199$, $SE = 0.450$, $t(775.28) = -4.884$, $p < .001$), indicating overall lower amplitudes in the ASD group.

Simple effects analyses within the TD group showed a clear grammaticality effect in posterior regions, with amplitudes for matching conditions significantly larger than for mismatching conditions (estimate = 1.232 μV, $SE = 0.406$, $t(774) = 3.034$, $p = .0025$). No significant effect was found in anterior regions ($p = .4915$). In contrast, the ASD group displayed an atypical pattern: in anterior regions, amplitudes for mismatching conditions were significantly larger than for matching conditions (estimate = –1.063 µV, SE = 0.300, $t(774) = -3.545$, $p = .0004$). In posterior regions, the grammaticality effect (match > mismatch) was only marginally significant ($p = .0684$). Between-group comparisons indicated that, in the posterior region under mismatching conditions, the difference between groups was not significant ($p = .1085$, Cohen's $d = 0.136$). Topographic maps (Figure 5) illustrated that TD children showed a typical posterior N400 effect, whereas ASD children exhibited a "reversed" effect in anterior regions and an attenuated effect in posterior regions.

N400 latency analysis was performed on difference waves at fronto-central electrodes. All 10 TD children showed a clear negative deflection and were included; in the ASD group, 6 out of 9 children met the inclusion criterion. The mean peak latency was 401.6 ms (SD = 69.0 ms) for the TD group and 380.0 ms (SD = 45.7 ms) for the ASD group. This difference was not statistically significant ($t(14) = 0.68$, $p = .509$; Cohen's $d = 0.35$).

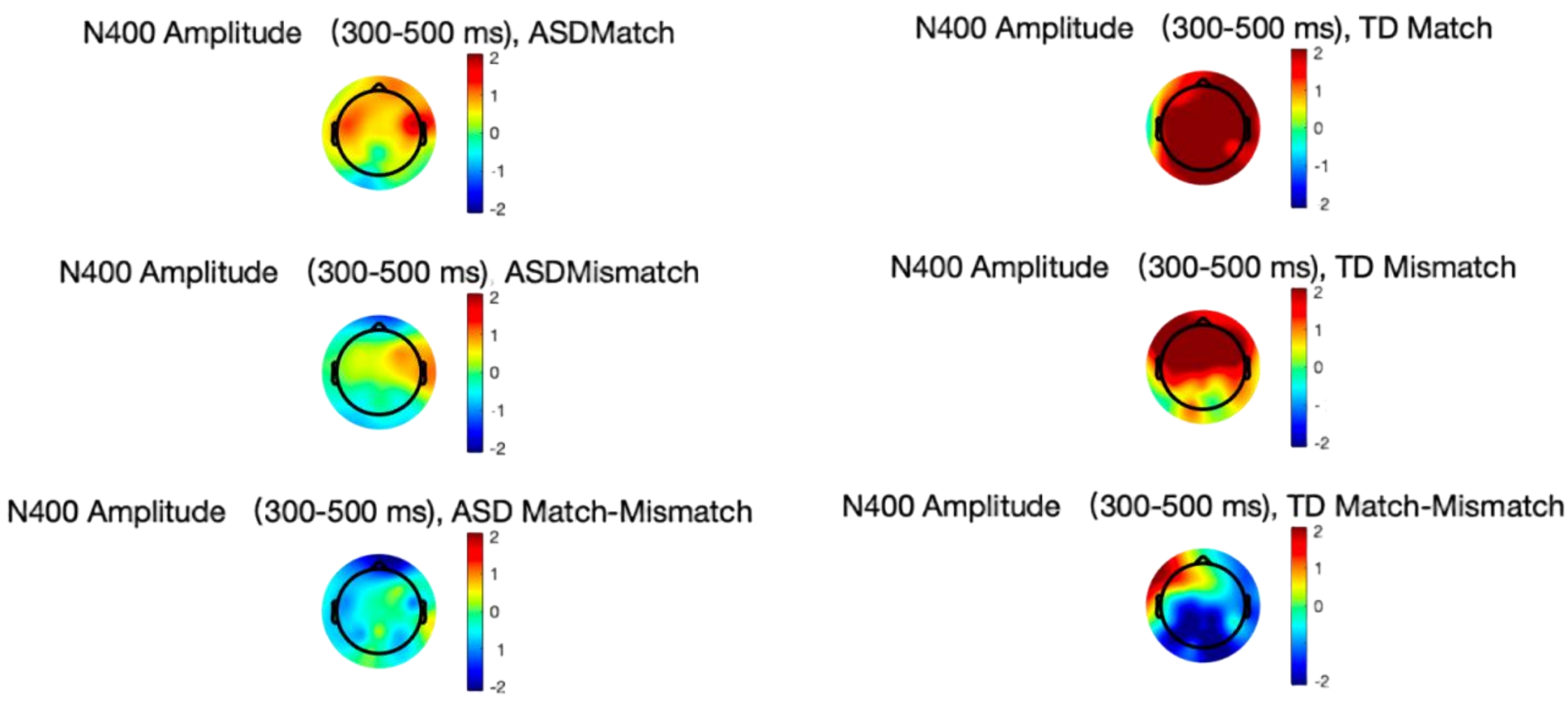

Figure 5. Topography (300ms-500ms) in ASD and TD group

**3.3 P600 (500–1000 ms)**

For the P600 component, the LMM revealed a significant main effect of Group ($\beta = -2.99$, SE = 0.43, $t(776.93) = -6.98$, $p < .001$), with ASD children showing overall more negative amplitudes than TD children. A significant Grammaticality × Anteriority interaction was found ($\beta = -2.12$, SE = 0.71, $t(774.08) = -3.00$, $p = .003$). Simple effects analysis showed that a significant grammaticality effect (match > mismatch) was present only in posterior regions (difference = 0.64 µV, SE = 0.26, $t(774) = 2.47$, $p = .014$), but not in anterior regions ($p = .217$). Within the TD group, this posterior grammaticality effect was robust (match vs. mismatch: difference = 1.23 μV, SE = 0.39, t(774) = 3.18, p = .002). In contrast, the ASD group showed no significant amplitude differences between matching and mismatching conditions in either anterior or posterior regions (all ps > .29). Topographic maps (Figure 6) and the three-way interaction plot (Figure 7) confirmed that the typical posterior P600 effect was absent in the ASD group.

P600 latency analysis, performed on difference waves at fronto-central electrodes, included all 10 TD children and 8 out of 9 ASD children. The mean peak latency was 678.4 ms (SD = 125.9 ms) for the TD group and 780.5 ms (SD = 156.5 ms) for the ASD group. Although the difference did not reach statistical significance (t(16) = –1.54, p = .144), the effect size was medium-to-large (Cohen's d = –0.73), suggesting a meaningful trend toward delayed syntactic processing in the ASD group.

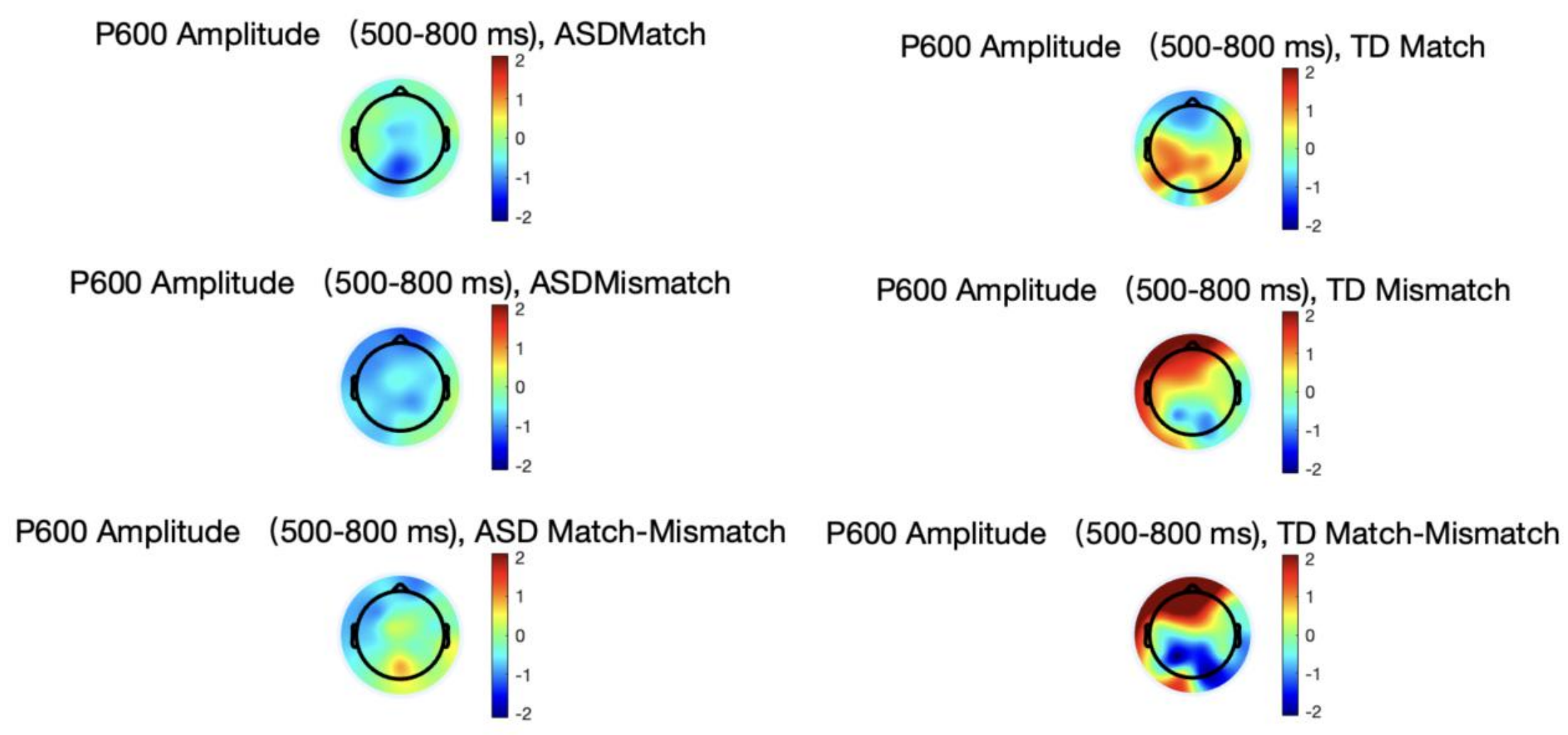

Figure 6. Topography (500ms-1000ms) in ASD and TD group

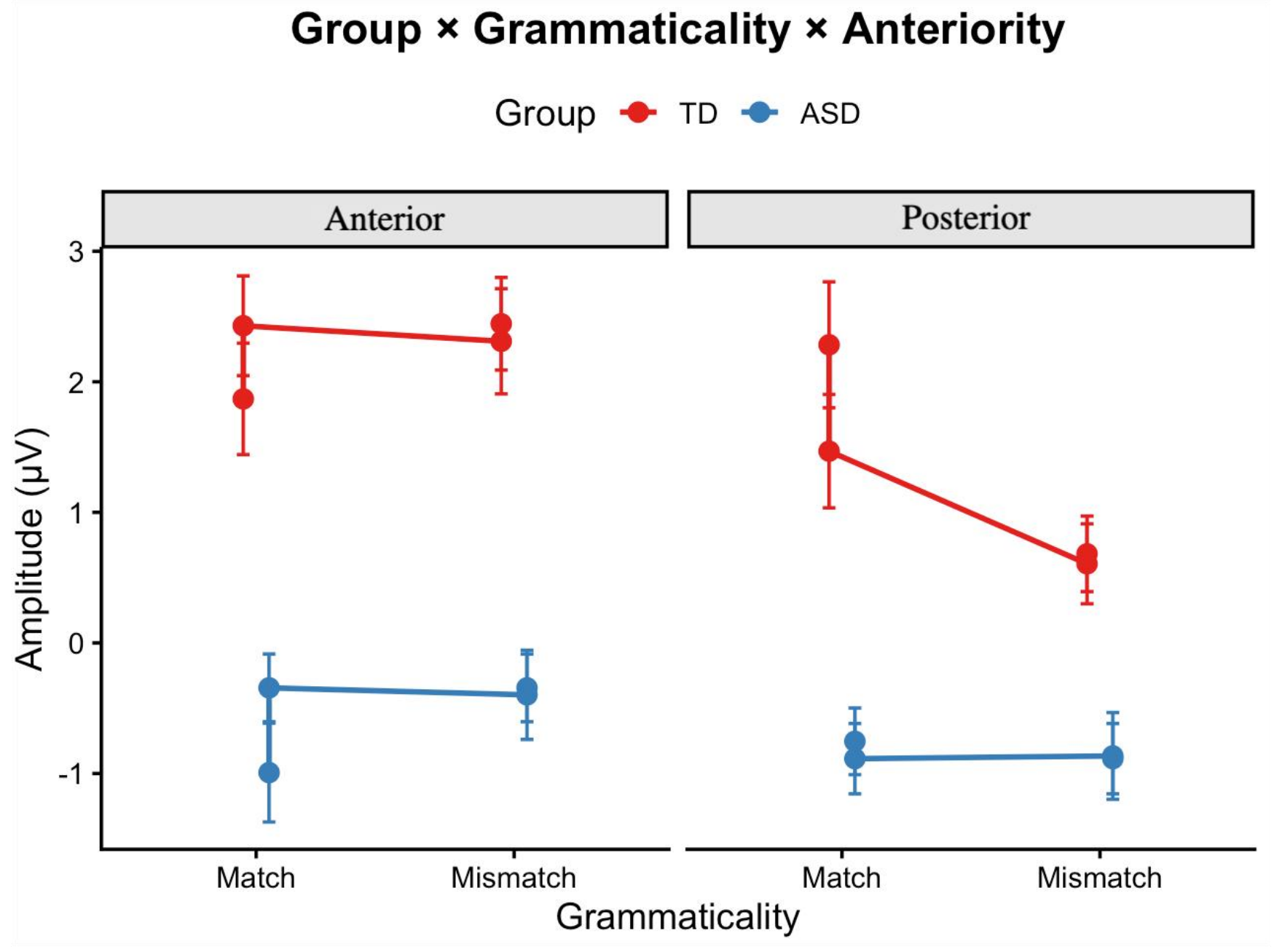

Figure 7. Three-way Interaction Plot of P600 (Group × Grammaticality × Anteriority)

Taken together, the ERP results revealed three core characteristics of recursive possessive structure processing in children with ASD. First, the P200 findings indicated atypical early perceptual processing: the ASD group showed significantly reduced overall amplitudes and failed to exhibit the typical posterior grammaticality effect observed in TD children. Second, the N400 results demonstrated that semantic integration was relatively preserved in the ASD group (no latency delay), but with an atypical spatial distribution—a reversed effect in anterior regions and an attenuated effect in posterior regions. Third, the P600 findings revealed a clear syntactic integration deficit in ASD children: they showed significantly reduced P600 amplitudes, no significant grammaticality effect in posterior regions, and a trend toward delayed latency. These patterns suggest that while fundamental semantic processing may be spared, the online computation and integration of hierarchical possessive relations are impaired in children with ASD, particularly at the stage of syntactic reanalysis.

**4. Discussion**

In this study, we investigated the online processing of two-level recursive possessive structures in Mandarin-speaking children with high-functioning autism spectrum disorder (ASD) compared to typically developing (TD) children, using event-related potentials (ERPs). By analyzing the P200, N400, and P600 components, we aimed to elucidate the cognitive neural mechanisms underlying recursive syntactic processing in ASD. Our findings revealed three core characteristics: atypical early perceptual processing (P200), relatively preserved but spatially atypical semantic integration (N400), and a significant deficit in syntactic reanalysis (P600). Together, these results provide neurophysiological evidence for a dissociation between semantic and syntactic processing in children with ASD, supporting theoretical accounts of modular impairment and compensatory mechanisms.

The P200 component, which indexes early perceptual and attentional processes (Cheng et al., 2025), showed no significant group difference in latency, indicating that basic perceptual timing may be preserved in ASD. However, the ASD group exhibited significantly lower overall P200 amplitudes and failed to show the typical posterior grammaticality effect (match > mismatch) observed in TD children. This pattern suggests that children with ASD allocate neural resources atypically during the earliest stages of syntactic structure processing. The absence of a region-specific P200 effect in posterior language-related areas (e.g., temporoparietal regions) is consistent with previous reports of reduced early sensitivity to structural violations in ASD (Pijnacker et al., 2010; Fishman et al., 2011). Importantly, the lack of group difference under mismatching conditions in posterior regions, combined with a significant difference under matching conditions, implies that the primary difficulty lies not in detecting violations per se, but in forming stable neural representations of expected grammatical structures. This early atypicality may have cascading effects on later syntactic integration (Marquez-Garcia et al., 2022), a possibility supported by the subsequent P600 findings.

The N400 results revealed a complex but informative pattern. Critically, no significant group difference in N400 latency was observed, and both groups showed comparable N400 effects in terms of presence, suggesting that lexical-semantic access and basic meaning integration are relatively preserved in high-functioning ASD. This aligns with previous ERP studies showing that individuals with ASD can successfully process semantic relations at the word level (Coderre et al., 2017; DiStefano et al., 2019). However, the spatial distribution of the N400 effect differed markedly between groups. TD children showed a typical posterior N400 effect (greater negativity for mismatching conditions), whereas ASD children exhibited a "reversed" effect in anterior regions (mismatch > match) and only a marginally significant effect in posterior regions. This atypical topography suggests that while semantic knowledge may be intact, the neural networks recruited for semantic integration are reorganized in ASD. The anterior "reversal" could reflect compensatory engagement of frontal regions to support semantic processing (Coderre et al., 2017). Importantly, the absence of a robust N400 violation effect in either group for the mismatch condition confirms that the critical manipulation (reversal of possessor order) was not perceived as a semantic anomaly. This finding is theoretically significant: it indicates that the processing difficulty in recursive possessives arises from syntactic hierarchical relations, not from lexical-semantic incongruity. Thus, the semantic pathway appears to be a relative strength in ASD, consistent with the "selective preservation" hypothesis (Boucher, 2012).

The most striking group differences emerged in the P600 component, which is widely associated with syntactic reanalysis, integration, and conflict monitoring (Friederici et al., 1993). While TD children exhibited a robust, posteriorly distributed P600 effect (higher amplitudes for matching than mismatching conditions), ASD children showed significantly reduced P600 amplitudes and no significant grammaticality effect in any brain region. Moreover, although the latency difference did not reach statistical significance, the medium-to-large effect size (Cohen's d = –0.73) suggests a meaningful trend toward delayed syntactic reanalysis in ASD. These findings provide direct neural evidence for a core deficit in online syntactic computation of recursive structures in children with ASD. The inability to generate a typical P600 response indicates that ASD children fail to detect the structural violation in recursive possessive sentences or to initiate the necessary reanalysis processes. This deficit is not due to a general inability to process language, as the same children showed relatively preserved N400 responses. Rather, it appears to be specific to the hierarchical integration required by recursion, supporting the "weak central coherence" theory (Happé & Frith, 2006), which posits that individuals with ASD struggle to integrate local elements into a global structure. Recursive possessives demand exactly such global hierarchical processing, and the absence of a P600 effect in the ASD group reflects this bottleneck.

An overarching theme of our results is the dissociation between the semantic and syntactic processing routes in children with ASD. In TD children, both routes operated in a collaborative manner: the N400 and P600 effects were both posteriorly distributed and showed consistent patterns. In ASD children, however, the semantic route was relatively preserved (normal N400 latency, though with atypical topography), whereas the syntactic route was severely impaired (reduced P600 amplitude, absent effect, delayed latency). This pattern suggests that ASD children may rely on semantic information to partially compensate for syntactic deficits, a strategy that has been observed behaviorally (Liu et al., 2019) and in ERP studies of semantic processing in ASD (Marquez-Garcia et al., 2022). However, such compensation has clear limits: semantic information alone cannot fully resolve the hierarchical relations encoded by recursive syntax. The failure to generate a P600 effect indicates that even when semantic processing is intact, the syntactic computation required for recursive structures remains unavailable or inefficient. This supports the "modular dissociation" perspective (Boucher, 2012; Coderre et al., 2017), which argues that different components of the language system can be selectively affected in ASD.

Our findings have several theoretical implications. First, they provide strong evidence against a domain-general semantic deficit in high-functioning ASD, instead supporting a more nuanced view in which syntactic processing is disproportionately impaired while semantic processing remains relatively preserved. Second, they demonstrate that recursion, as a core property of human language, is computationally challenging for children with ASD, even when their verbal IQ is within the normal range. This challenges the notion that language difficulties in ASD are primarily pragmatic or lexical, highlighting the need for targeted syntactic interventions. Clinically, the attenuated P600 and delayed latency could serve as neural markers for assessing syntactic processing efficiency in ASD. Intervention programs should focus on explicit training of hierarchical structure building, perhaps using visual supports (e.g., color-coded possessor chains) to reduce cognitive load and facilitate syntactic integration.

Several limitations should be acknowledged. The sample size was modest, and the ASD group was restricted to high-functioning children, limiting generalizability to the broader autism spectrum. The cross-sectional design precludes conclusions about developmental trajectories. Additionally, we examined only one type of recursive structure (two-level possessives); future studies should include higher levels of embedding and other recursive forms (e.g., relative clauses) to determine the generality of the observed deficits. Finally, the absence of a significant N400 effect in either group for the mismatch condition, while theoretically informative, should be replicated with a larger sample and a more sensitive semantic manipulation. Future research should also employ source localization and functional connectivity analyses to identify the specific brain regions and networks underlying the atypical P200 and P600 patterns observed in ASD.

Several limitations of the present study should be acknowledged, which also point to promising avenues for future research.

First, the relatively modest sample size constitutes a significant limitation. Despite rigorous screening procedures, the final analysis included only 12 children with ASD and 12 TD children. This sample size, while consistent with many previous ERP studies on special populations (Megnin et al., 2012), limits statistical power and the generalizability of the findings. The inherent variability in electrophysiological data, combined with the heterogeneity of the autism spectrum, increases the risk of Type II errors. Future research should therefore prioritize larger sample sizes, ideally through multi-center collaborations that can pool resources and recruit more diverse participant cohorts. Such collaborative efforts would not only enhance statistical power but also improve the representativeness of findings across different clinical settings and geographic regions.

Second, the heterogeneity inherent to the autism spectrum poses fundamental challenges. Although systematic screening procedures were implemented (DSM-5 criteria, PPVT, WISC-CR), the participating ASD children inevitably varied along multiple dimensions, including cognitive profiles, language abilities, symptom severity, and co-occurring conditions. This heterogeneity, while reflective of the true nature of ASD (Bu & Xu, 2015), introduces considerable variability into the data and complicates the interpretation of group-level findings. Future studies should adopt approaches that explicitly address this heterogeneity, such as larger-scale studies that stratify participants based on relevant dimensions (e.g., language profile, cognitive level, age). More fine-grained characterization of participants' language abilities beyond global measures—including standardized assessments of syntactic comprehension, working memory capacity (Liu et al., 2019), and processing speed—would allow for more precise examination of the relationships between specific cognitive-linguistic skills and recursive processing outcomes.

Third, the scope of behavioral data collected was relatively circumscribed. While the sentence-picture matching paradigm successfully recorded accuracy and reaction time measures, the binary judgment task provides limited insight into the nature of children's interpretations or the specific points at which processing difficulties arise. When children responded incorrectly or slowly, the current data do not reveal whether errors stemmed from failure to parse the recursive structure, confusion about the possessive relations depicted in the pictures, memory limitations, attentional fluctuations, or decision-level difficulties. Future research should therefore incorporate richer behavioral assessments that complement online ERP measures. Post-experiment comprehension questions could probe children's explicit understanding of the possessive relations, following truth-value judgment paradigms (Chien & Chen, 2024). Eye-tracking measures combined with ERP recordings could provide additional insights into the time course of visual exploration and its relationship to linguistic processing (Ding et al., 2015).

Fourth, the study focused on a single type of recursive structure: two-level recursive possessives. While this structure represents an important test case for investigating hierarchical processing (Terunuma et al., 2017; Shi et al., 2019; Li et al., 2020), the findings may not generalize to other recursive forms such as recursive relative clauses, recursive prepositional phrases, or recursive complement constructions, which may impose different processing demands and engage partially different neural mechanisms. Different types of recursive structures may follow distinct acquisition trajectories even within the same population (Pérez-Leroux et al., 2012, 2022). Moreover, the focus on two-level recursion leaves open questions about how children with ASD would process structures with higher levels of embedding, which would place greater demands on working memory and computational resources (Zhang & Zan, 2007). Future research should extend the investigation to a broader range of recursive structures, systematically manipulating the type of recursion and the level of embedding to map the boundaries of recursive processing abilities in ASD.

In summary, while the present study provides novel insights into the neural mechanisms of recursive processing in children with ASD, these limitations highlight the need for larger-scale, longitudinal, and multi-modal investigations. Addressing these issues in future research will not only deepen our theoretical understanding of language in autism but also inform the development of more effective, individualized intervention strategies.

**Author contribution statement:**

(Please note that one author may have more than one role or a job may have been done by multiple authors)

Fu Chenxi: making the EEG materials; collecting the EEG data; writing the paper.

Wang Xiaoyi: collecting the EEG data

Ziman Zhuang: collecting the EEG data

Yang Caimei: supervising the whole EEG research group working at a series of theoretical and experimental studies related to recursion, part of which covers the present paper; designing the whole research project; providing financial support for the project.

**Funding acknowledgement: T**his work was supported by the National Social Science Fund of China (Grant Number: 23BYY170)